\documentclass[conference,letterpaper]{IEEEtran}
\usepackage[letterpaper, left=0.7in, right=0.7in, bottom=0.99in, top=0.7in]{geometry}
\IEEEoverridecommandlockouts

\usepackage[font=footnotesize,caption=false]{subfig} 

\usepackage{cite}
\usepackage{amsmath,amssymb,amsfonts, mathtools, amsthm}
\usepackage{algorithmic}
\usepackage{graphicx}
\usepackage{textcomp}
\usepackage{xcolor}
\usepackage{booktabs}
\usepackage{glossaries}
\usepackage{subfig}
\usepackage{algorithm}
\usepackage{algorithmic}
\usepackage{bbm}
\usepackage{bm}
\usepackage{multirow}

\usepackage[normalem]{ulem}

\usepackage{balance}

\newacronym{bcd}{BCD}{Block Coordinate Descent}
\newacronym{leo}{LEO}{Low Earth Orbit}
\newacronym{isl}{ISL}{Inter-Satellite Link}
\newacronym{gsd}{GSD}{ground sample distance}
\newacronym{fov}{FoV}{field of view}
\newacronym{gtfp}{GTFP}{ground track frame period}
\newacronym{gs}{GS}{ground station}
\newacronym{fso}{FSO}{free-space optical}
\newacronym{smec}{SMEC}{satellite mobile edge computing}
\newacronym{mec}{MEC}{mobile edge computing}
\newacronym{pdd}{PDD}{Penalty Dual Decomposition}
\newacronym{aodv}{AODV}{Ad hoc On-demand Distance Vector}
\newacronym{cnn}{CNN}{convolutional neural networks}
\newacronym{dl}{DL}{Deep Learning}
\newacronym{dod}{DoD}{depth of discharge}
\newacronym{dqn}{DQN}{deep Q-learning}
\newacronym{gsl}{GSL}{ground-to-satellite link}
\newacronym{bm}{BM}{benchmark}
\newacronym{ml}{ML}{Machine Learning}
\newacronym{mdp}{MDP}{Markov decision process}
\newacronym{ngeo}{NGEO}{Non-geostationary orbit}
\newacronym{olsr}{OLSR}{optimized link state routing protocol}
\newacronym{ospf}{OSPF}{Open Shortest Path First}
\newacronym{pan}{PAN}{Path-Aware Networking}
\newacronym{qos}{QoS}{Quality of Service}
\newacronym{rl}{RL}{Reinforcement Learning}
\newacronym{drl}{DRL}{Deep \gls{rl}}
\newacronym{dnn}{DNN}{Deep Neural Network}
\newacronym{dql}{DQL}{Deep Q-learning}
\newacronym{e2e}{E2E}{end-to-end}
\newacronym{bgp}{BGP}{Border Gateway Protocol}
\newacronym{ibgp}{iBGP}{interior Border Gateway Protocol}
\newacronym{ebgp}{eBGP}{exterior Border Gateway Protocol}
\newacronym{as}{AS}{Autonomous System}
\newacronym{relu}{ReLu}{Rectified Linear Unit}
\newacronym{cdf}{CDF}{Cumulative Distribution Function}
\newacronym{ntn}{NTN}{Non-Terrestrial Networks}
\newacronym{lsatc}{LSatC}{\gls{leo} Satellite Constellation}
\newacronym{ai}{AI}{Artifical Intelligence}
\newacronym{ip}{IP}{Internet Protocol}
\newacronym{ue}{UE}{User Equipment}
\newacronym{pomdp}{POMDP}{Partially Observable Markov Decision Problem}
\newacronym{hol}{HOL}{Head of Line}
\newacronym{fifo}{FIFO}{First-In First-Out}
\newacronym{snr}{SNR}{Signal-to-Noise Ratio}

\def\BibTeX{{\rm B\kern-.05em{\sc i\kern-.025em b}\kern-.08em
    T\kern-.1667em\lower.7ex\hbox{E}\kern-.125emX}}

\title{Multi-Agent Deep Reinforcement Learning for Distributed Satellite Routing\vspace{-0.2em}}

\author{\IEEEauthorblockN{Federico Lozano-Cuadra, Beatriz Soret~\IEEEmembership{Senior Member,~IEEE}\vspace{-0.5em}}

\thanks{F. Lozano-Cuadra (flozano@ic.uma.es) and B. Soret are with the Telecommunications Research Institute, University of Malaga, 29071, Malaga, Spain. This work is partially funded by ESA SatNEx V (prime contract no. 4000130962/20/NL/NL/FE), and by the Spanish Ministerio de Ciencia, Innovación y Universidades (PID2022-136269OB-I00). 
}
}

\date{}

\def\subparagraph{} 
\usepackage{titlesec}
\titlespacing*{\section}{0pt}{*1}{*1}
\titlespacing{\subsection}{0pt}{*1}{*1}

\renewcommand{\thesubsubsection}{\arabic{subsubsection}}

\titleformat{\subsubsection}[runin]{\itshape}{\thesubsubsection)}{1em}{}
\titlespacing*{\subsubsection}{\parindent}{0pt}{*1}

\begin{document}

\bstctlcite{IEEEexample:BSTcontrol}

\maketitle
\begin{abstract}

This paper introduces a Multi-Agent Deep Reinforcement Learning (MA-DRL) approach for routing in Low Earth Orbit Satellite Constellations (LSatCs). Each satellite is an independent decision-making agent with a partial knowledge of the environment, and supported by feedback received from the nearby agents. Building on our previous work that introduced a Q-routing solution, the contribution of this paper is to extend it to a deep learning framework able to quickly adapt to the network and traffic changes, and based on two phases: (1) An offline exploration learning phase that relies on a global Deep Neural Network (DNN) to learn the optimal paths at each possible position and congestion level; (2) An online exploitation phase with local, on-board, pre-trained DNNs. Results show that MA-DRL efficiently learns optimal routes offline that are then loaded for an efficient distributed routing online.


\end{abstract}
\vspace{-.1cm}
\glsresetall

\section{Introduction}

Low Earth Orbit (LEO) Satellite Constellations (LSatCs) are one of the pillars of 6G ubiquitous and global connectivity, enhancing cellular coverage, supporting a global backbone, and enabling advanced applications~\cite{Leyva-Mayorga2020}. Unlike terrestrial networks with stable links that can be handled with Dijkstra's algorithm and static routing tables, the unique characteristics of LSatCs calls for specific routing solutions. 
Specifically, LSatCs deal with rapidly moving satellites, predictable yet dynamic topology, significant propagation delays, and unbalanced and unpredictable terrestrial traffic~\cite{Rabjerg2021}. 

\begin{figure}[t]
    \centering
    {\includegraphics[width=3.5in]{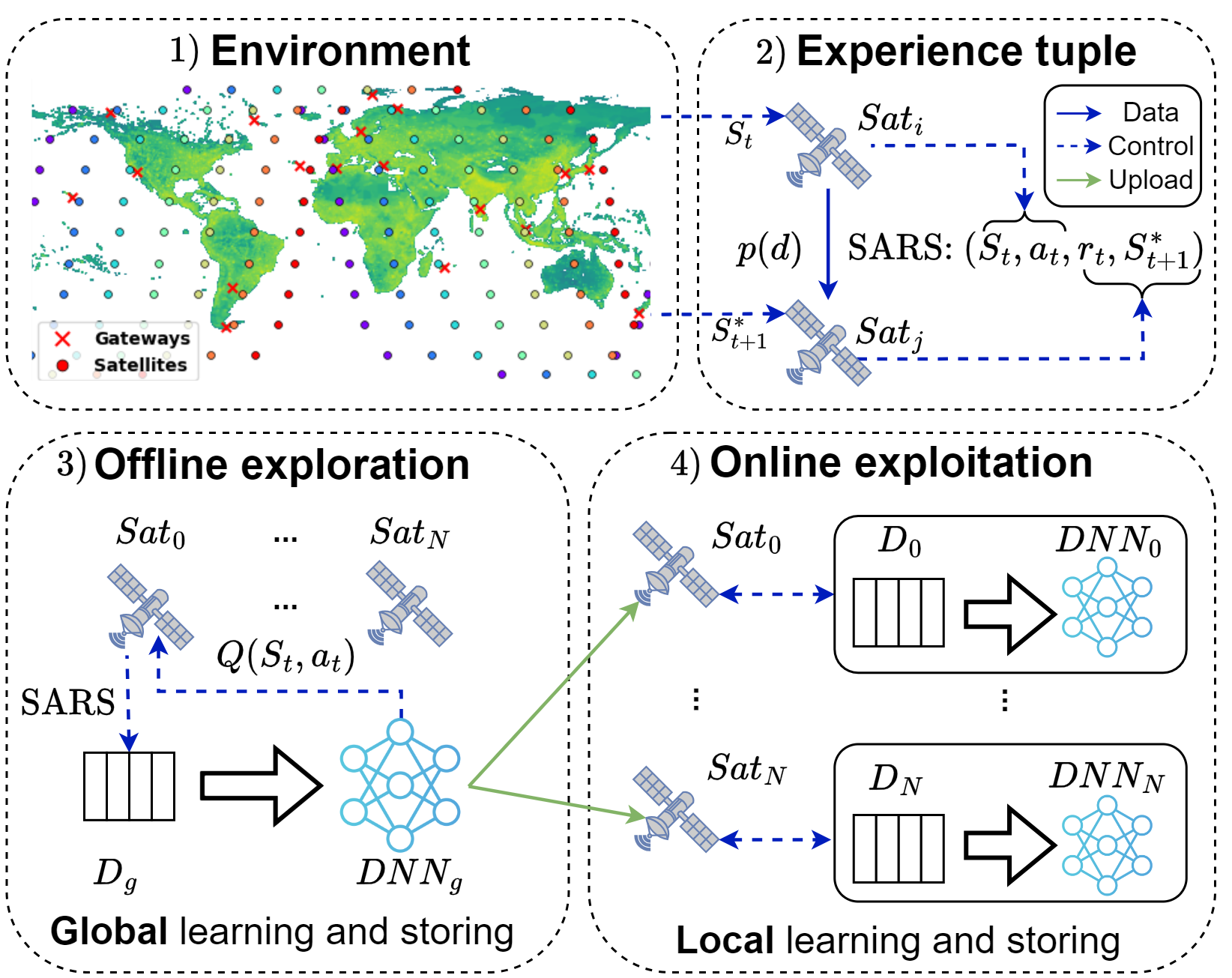}}
    \caption{System model: 1) network graph; 2) multi-agent interaction needed to build the tuple experience ($\text{SARS}$); 3) and 4) are representations of the offline exploration and online exploitation phases, respectively.}
    \label{fig:Phases}\vspace{-0.5cm}
\end{figure}

This paper introduces a novel approach for the End-to-End (E2E) packet routing in LSatCs, avoiding the dependence upon the ground infrastructure and aiming for a robust, low-latency solution. Building on our previous work in Q-routing~\cite{soret2023q}, we extend it to a deep learning framework able to handle a more complex state space, including local position and congestion information. This allows the agent to adapt easily to new situations. Our approach utilizes Multi-Agent Deep Reinforcement Learning (MA-DRL) where each satellite acts as a different agent with partial knowledge of the environment, informed by feedback from nearby agents. Unlike previous Machine Learning applications in routing, which have struggled with dynamic queuing times and multi-agent interactions~\cite{fadlullah2017deep, Liu2021, Liu2022}, our MA-DRL algorithm incorporates a two-phase solution: (1) An offline exploration learning phase  utilizing a global Deep Neural Network ($DNN_g$) to learn optimal paths for each position and congestion condition (Fig. \ref{fig:Phases}.3); (2) An online exploitation phase with local, on-board, pre-trained DNNs (Fig. \ref{fig:Phases}.4). This paper presents a comprehensive model of LSatC and ground infrastructure, formulating routing as a Partially Observable Markov Decision Problem (POMDP)~\cite{soret2023q}, and demonstrates that our MA-DRL algorithm learns and utilizes optimal routes for distributed routing.

\section{System Model} \label{sec:systemmodel}

The LSatC network, formed by both space and ground layers, is abstracted as a graph. The \textbf{space segment} consists of $N$ satellites across $M$ orbital planes (Fig. \ref{fig:Phases}.1), forming a finite set of satellite nodes, $\mathbb{S}$, and a set of edges, $\mathcal{E}$, representing the transmission links between them. Each satellite $Sat_i$ is equipped with 2 antennas for intra-plane and 2 for inter-plane communication, with their feasible edge set, $\mathcal{E}_i$.

The \textbf{ground segment} includes a set of gateways (Fig. \ref{fig:Phases}.1), $\mathbb{G}$, located at key global positions (Fig. \ref{fig:Phases}.1). These gateways maintain a single ground-to-satellite link (GSL) with their nearest satellite, constituting the edge set $\mathcal{E}_{G}$.
The data rate for communication is determined by the highest modulation and coding scheme possible within the current Signal-to-Noise Ratio (SNR), in line with DVB-S2 standards. 
Each gateway gathers the ground traffic and distributes it to each other gateway equally by injecting it to the LSatC.

The latency is computed considering the queue time at the satellite, transmission time based on data rate and propagation time over the link distance. This latency model accounts for varying traffic loads, where propagation time dominates in non-congested networks, but queue time quickly escalates under high traffic conditions~\cite{Rabjerg2021}.


\textbf{Benchmarking} involves comparing our MA-DRL algorithm with a traditional shortest path routing approach using Dijkstra's algorithm, where edge weights are proportional to the slant range between nodes. %

\section{Learning framework} \label{sec:learning}
In our MA-DRL, each satellite works as an independent agent in a networked multi-agent system, where the decision-making process for routing data packets is based on a POMDP. Upon packet arrival, each agent observes the \textbf{state} $S_t$ and makes an \textbf{action} $a_t$. The observed $S_t$ includes information about the agent's position, neighboring agents positions, packet destination and neighboring agents congestion levels. The $a_t$ to take consists on selecting the next hop for forwarding the packet. Afterwards, the \textbf{reward} $r_t$ for the $(S_t, a_t)$ pair is based slant range reduction from the packet to its destination after being forwarded and time spent on the receiving agent queue.

In conventional DRL, an agent $i$ stores every tuple of experiences $(S_t, a_t, r_t, S_{t+1})$ in order to learn, where $S_{t+1}$ is the state where $i$ has transited to and $r_t$ is the reward after taking $a_t$ at the observed state $S_t$. The innovative aspect of the MA-DRL algorithm lies in observing the impact of an action $a_t$ from the perspective of a packet $p$ with destination $d$, $p(d)$. When $p(d)$ is forwarded by an agent $Sat_i$ to another agent $Sat_j$, it transits from the state $S_t$ observed in $Sat_i$ to $S^*_{t+1}$ observed at $Sat_j$. This experience tuple $\text{SARS:}$ $(S_t, a_t, r_t, S^*_{t+1})$ with states observed in the interacting agents is then stored in a experience buffer $D$ and used to train a DNN that learns the optimal routing policy (Fig.~\ref{fig:Phases}.2).

The learning process involves two phases. The first is the \textbf{offline exploration} (Fig.~\ref{fig:Phases}.3), where there is a global experience buffer $D_g$ where experiences from all the agents are stored and used to train a global $DNN_g$. After $DNN_g$ learns the optimal routing policy at every observed state the \textbf{online exploitation} phase starts (Fig.~\ref{fig:Phases}.4). Here every agent $i$ has its own $DNN_i$ onboard, which is a copy of the trained $DNN_g$ that dictates the routing policy. In both phases there is a minimal feedback needed between satellites where each agent $i$ needs its neighboring agents congestion information in order to observe $S_t$. Moreover, each agent $i$ needs the new state $S^*_{t+1}$ encountered at $p(d)$´s receiving agent $j$ and the time spent on its queue in order to compute $r_t$ (Fig \ref{fig:Phases}.2). This information is then stored in the local experience buffer $D_i$ in order let $i$ keep training $DNN_i$ with local data.

\section{Preliminary results} \label{sec:results}

\begin{figure}[t]
    \centering
    {\includegraphics[width=3.5in]{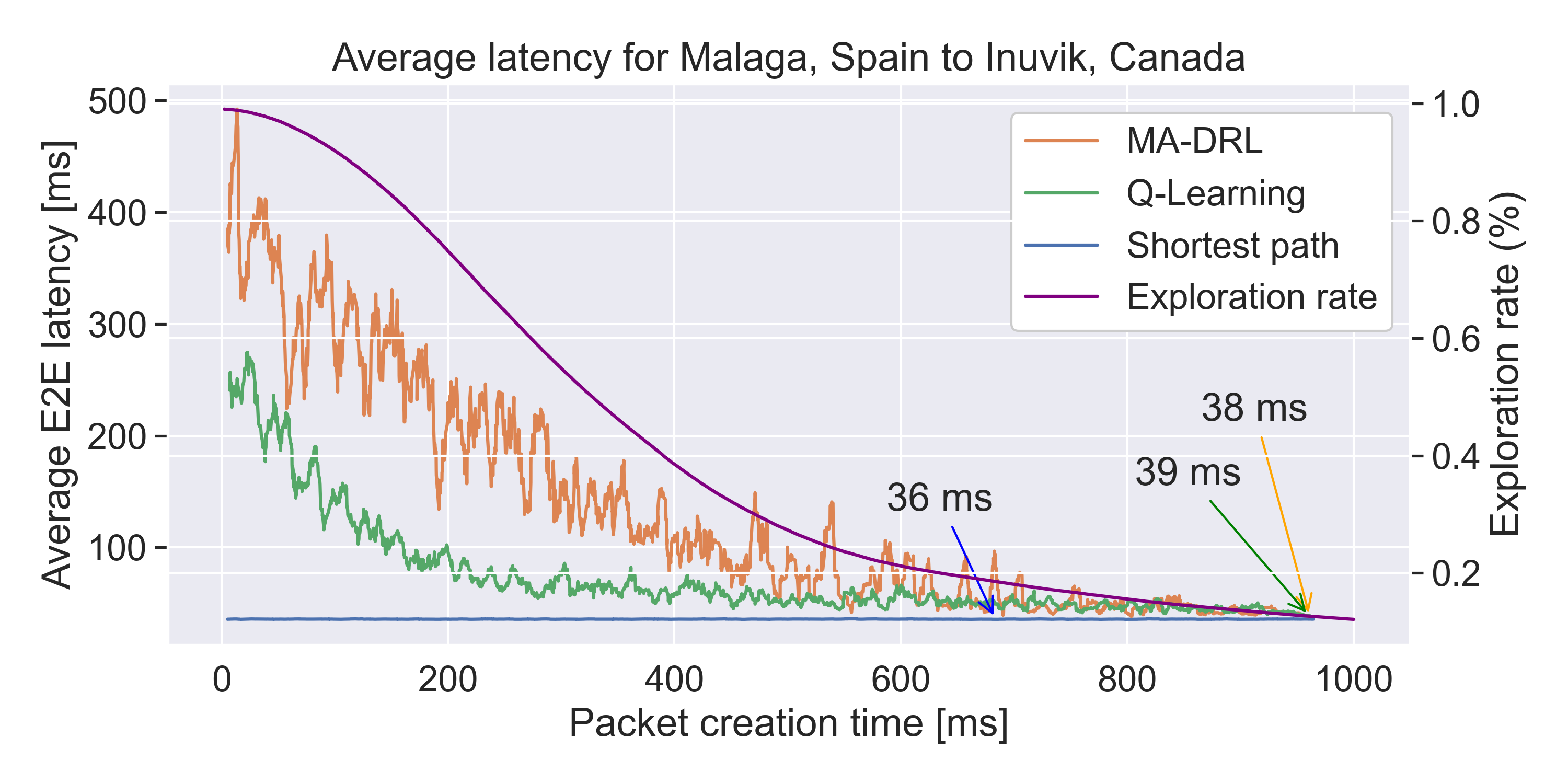}}
    \caption{E2E latency versus packet creation time. Comparison of our MA-DRL with the Q-Routing algorithm \cite{soret2023q} and the slant range shortest path benchmark.}
    \label{fig:latency}\vspace{-0.5cm}
\end{figure}

During the exploration phase, the weights $\theta_g$ that parameterize $DNN_g$ are initialized randomly. Coupled with a high exploration rate $\epsilon$, $DNN_g$ tends to make random routing actions initially. Fig. \ref{fig:latency} shows how this behaviour disappears as $\epsilon$ decreases: $DNN_g$ learns first sub-optimal paths and then converges to the shortest path in less than 1 second of real time simulated. 


Note that the Q-learning method \cite{soret2023q} converges faster than DNN due to simpler Q-Tables without positional data and reduced queue details, but it must continually converge to new solutions, which limits its adaptability to congestion and location changes, unlike MA-DRL.
The shortest path algorithm has real time information about all the LSatC, while MA-DRL has only 1 hop neighboring information at every hop, which is a more realistic approach; it is impractical to have real time information about the whole LSatC due to the congestion caused by feedback messages and propagation times delays. 


\section{Conclusions}
\label{sec:conclusions}


The implementation of MA-DRL in LSatC demonstrates promising results in terms of efficient learning, as it quickly converges to an optimal routing policy with minimal LSatC local status information at every hop, showcasing the effectiveness of the decentralized DNN-based approach. Notably, in terms of adaptability, MA-DRL has not only learned the optimal path but also a set of alternative paths during the offline exploration phase. This aspect becomes particularly advantageous in more loaded LSatCs where the agents, being aware of their neighbors' congestion status, can seamlessly switch to these alternative paths. This ability to adapt to changing network conditions by selecting appropriate routes underscores the robustness and practical utility of our MA-DRL approach in dynamic satellite environments.


\bibliographystyle{IEEEtran}
\bibliography{ref}
\end{document}